# Gleanmer: A 6 mW SoC for Real-Time 3D Gaussian Occupancy Mapping

Zih-Sing Fu[*], Peter Zhi Xuan Li[*], Sertac Karaman, Vivienne Sze
Massachusetts Institute of Technology, Cambridge, MA, USA
[*]These authors contributed equally to this work.

**Abstract**
High-fidelity 3D occupancy mapping is essential for many edge-based applications (such as AR/VR and autonomous navigation) but is limited by power constraints. We present Gleanmer, a system on chip (SoC) with an accelerator for GMMap, a 3D occupancy map using Gaussians. Through algorithm-hardware co-optimizations for direct computation and efficient reuse of these compact Gaussians, Gleanmer reduces construction and query energy by up to 63% and 81%, respectively. Approximate computation on Gaussians reduces accelerator area by 38%. Using 16nm CMOS, Gleanmer processes 640×480 images in real time beyond 88 fps during map construction and processes over 540K coordinates per second during map query. To our knowledge, Gleanmer is the first fabricated SoC to achieve real-time 3D occupancy mapping under 6 mW for edge-based applications.

## Introduction

3D occupancy maps classify the 3D environment into occupied, free, and unexplored regions using probabilities. While sparse maps (*e.g.*, LiDAR/SLAM point clouds [1]) only contain a few points within occupied regions of the environment, occupancy maps completely capture all spatial regions (see Fig. 1, top). This enables edge devices and their users to navigate safely by staying within free regions and avoiding high-risk occupied and unexplored regions.

Traditional 3D occupancy maps, such as OctoMap [2], lack compactness and consume over 100 mW even with hardware acceleration [3], exceeding edge power budgets [4][5][6]. To achieve superior energy efficiency and compactness, GMMap utilizes occupied (red) and free (blue) Gaussian ellipsoids (see Fig. 2) to represent occupied and free regions [7], respectively. These 3D Gaussians, defined by their means and covariances, are indexed via an R-Tree of their bounding boxes for efficient retrieval. The map is constructed frame-by-frame from sensor rays derived from depth images and their corresponding camera poses. *Gaussian Generation* extracts local occupied Gaussians from sensor endpoints and free Gaussian bases from ray paths of the current image. During *Gaussian Fusion*, these bases are refined into local free Gaussians; subsequently, all local Gaussians are fused into the global map. After construction, the map is queried via *Gaussian Regression*, which computes the occupancy probability at a given coordinate by evaluating its spatially overlapping Gaussians.

Existing embedded software implementations of GMMap consume over 2 W [7] and are prohibitive for many edge devices. Thus, we present Gleanmer to enable real-time mapping under 6 mW. Gleanmer leverages direct Gaussian computation and cross-query Gaussian reuse to reduce energy, while the silicon area is reduced via approximate computing.

## Architecture Overview

Fig. 3 shows Gleanmer's architecture with a GMMap accelerator. The CPU is used to reconfigure the accelerator. Depth images are streamed through external IOs into the Depth Decoder. Dedicated engines accelerate Gaussian Generation and Fusion. To ensure efficient map sharing without duplication, the constructed global map is stored in a global buffer accessible via a shared AXI-4 bus. The Gaussian Management Engine reduces map access latency by using an R-Tree Engine to organize Gaussians into a tree of bounding boxes, enabling Gaussian insertion, removal, and search in logarithmic time. Since these operations access one small subregion of the global map at a time, a 44 KB cache exploits the temporal locality of this subregion to increase construction throughput by 4 to 7× and query throughput by 3 to 9×.

## Energy-Efficient Free Gaussian Bases Generation

To generate local occupied Gaussians from a depth image, Scanline Segmentation (SS) first extracts line segments from image rows, which are then merged across adjacent rows via Segment Fusion (SF) to represent individual obstacles. Traditionally, free Gaussian bases are generated from the sensor rays associated with line segments, as shown in Fig. 4a. However, the computational cost scales with the total number of line segments (in the thousands) and accounts for 26% to 65% of the total map construction energy. To reduce energy usage by 22% to 63% (see Fig. 7a), the Free Gaussian Bases Generation Unit (see Fig. 3) computes free Gaussian bases *directly* from occupied Gaussians (see Fig. 4b). Since sensor rays are not stored during SS for memory efficiency, a small set of representative rays is sampled from each occupied Gaussian to efficiently generate these free Gaussian bases.

## Reusing Gaussians Across Successive Queries

To ensure collision-free motion, the device queries occupancy probabilities at coordinates along piecewise-linear trajectories in the constructed global map. Since trajectory coordinates are spatially proximate, they typically traverse similar R-Tree paths (see Fig. 5). To eliminate redundant path traversals, Gleanmer utilizes batch querying, where a single bounding box enclosing successive coordinates retrieves overlapping Gaussians for all these coordinates through the R-Tree. These Gaussians are time-interleaved across all coordinates to share the Gaussian Regression Unit (see Fig. 3) with a negligible 2% accelerator area overhead for registers storing intermediate results. This batching strategy for 16 coordinates increases query throughput by 4 to 10× while reducing energy consumption by 74% to 81% (see Fig. 7b).

## Reducing Area with Approximate Computation

The buffer area in Gaussian Generation is reduced by 8× by 1) generating free Gaussian bases directly from occupied Gaussians to avoid storing these bases for line segments, and 2) a single-cycle slope approximation in the feedback stage of the SS Unit so that line segments from only one depth-image row (instead of four) are buffered. The SS Unit uses a feedback stage to compute the line segment slope, which typically requires four cycles. To sustain throughput, the core must time-interleave across four rows, which requires storing 4× line segments. By approximating the slope in one cycle with its value from four cycles prior (*i.e.*, minimal slope variance across neighboring pixels of the same line in Stage 4 of Fig. 6), the SS Unit achieves equivalent throughput without time-interleaving. Finally, reducing Gaussian precision from 32-bit to 19-bit yields a 38% reduction in Fusion and Regression Engine area. Covariance matrices remain in 32-bit to prevent degeneracy. In total, these approximations reduce the accelerator area by 38% (Fig. 7c) and shrink the map size by 44 to 63% while maintaining map accuracy (Fig. 8, top).

## Implementation and Results

Fig. 8 summarizes Gleanmer's performance across three diverse environments [8][9] (see Fig. 9). Fig. 10 shows the chip testing setup and micrograph. With comparable accuracy, Fig. 11 shows that Gleanmer achieves higher construction and query throughput than the NVIDIA Jetson TX2 for GMMap [7] and the OMU accelerator for OctoMap [3]. For construction, Gleanmer processes 640×480 depth images in real-time above 88 fps. For query, Gleanmer achieves superior throughput of 540K to 1320K coordinates per second (cps). The power consumption is under 6 mW which is at least 341× lower than NVIDIA Jetson TX2 and 44× lower than OMU.

**Acknowledgements:** MIT-MathWorks Fellowship, Amazon, NSF CPS 2400541, Intel USP for silicon donation.

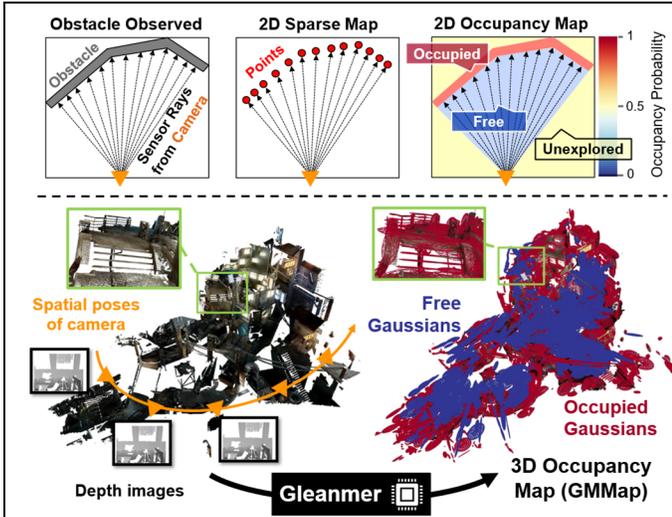

Fig. 1. Top: Visual differences between a 2D sparse map from point cloud and a 2D occupancy map. Bottom: Gleanmer uses a sequence of depth images and their camera poses to construct a 3D occupancy map (GMMap) consisting of occupied and free Gaussians.

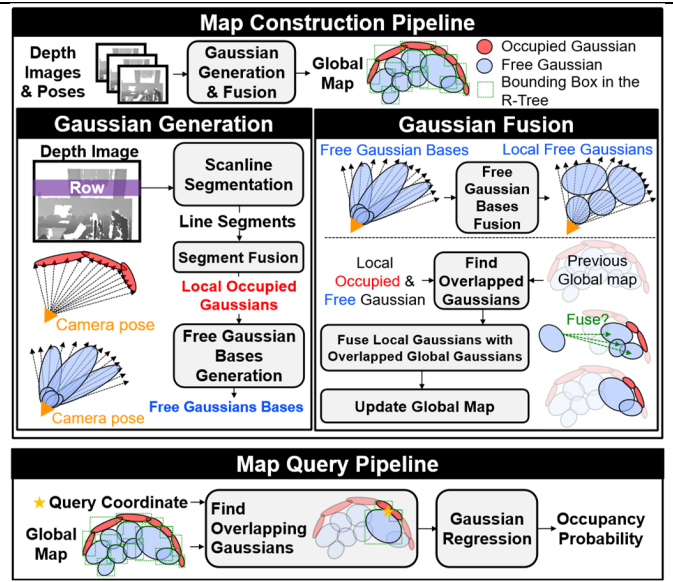

Fig. 2. Simplified illustration of GMMap for map construction (Gaussian Generation and Fusion) and query (Gaussian Regression).

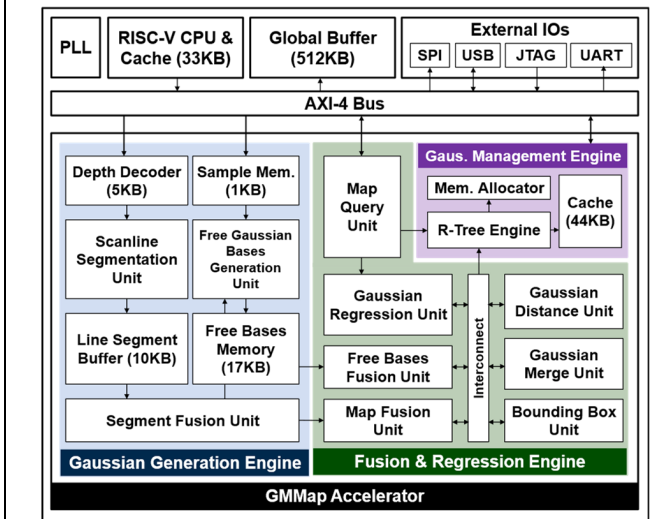

Fig. 3. System architecture of Gleanmer SoC, including a RISC-V CPU, 512 KB global buffer, external I/Os, and a GMMap accelerator. The GMMap accelerator is composed of dedicated engines to accelerate Gaussian Generation, Fusion and Regression.

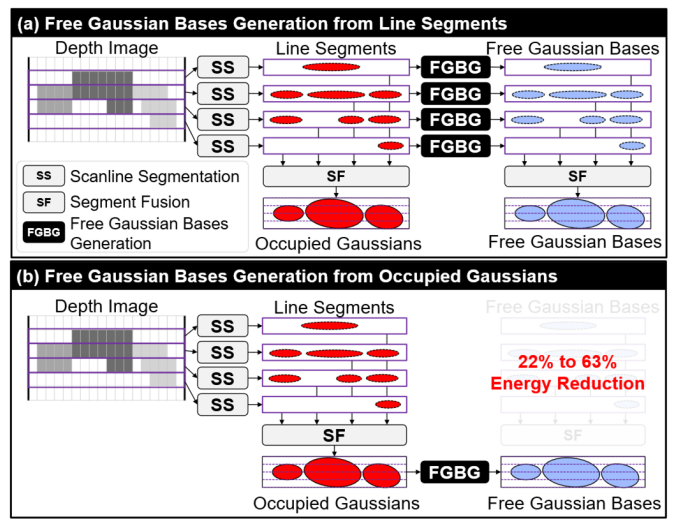

Fig. 4. (a) An example of generating free Gaussian bases from line segments, which dominates up to 65% of map construction power. (b) To reduce energy by 22% to 63%, free Gaussian bases are directly from occupied Gaussians instead.

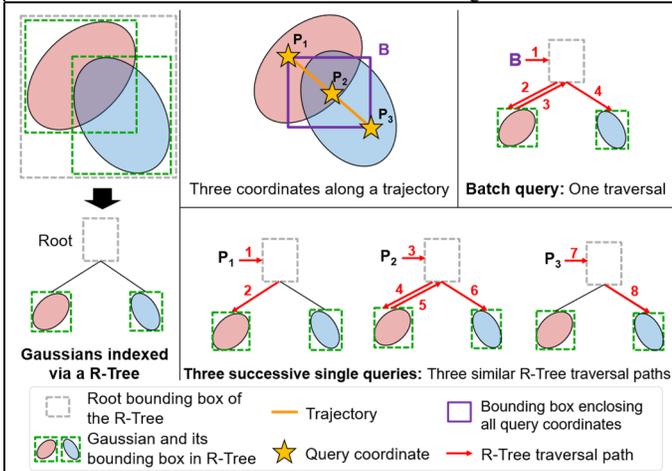

Fig. 5. Comparison between single and batch query at three successive coordinates along a trajectory. For this example, querying one coordinate at a time (single query) results in three similar R-Tree traversal paths with a total length of 8. To reduce memory accesses associated with traversal, querying a batch of coordinates using an enclosing bounding box (batch query) reduces path length to 4.

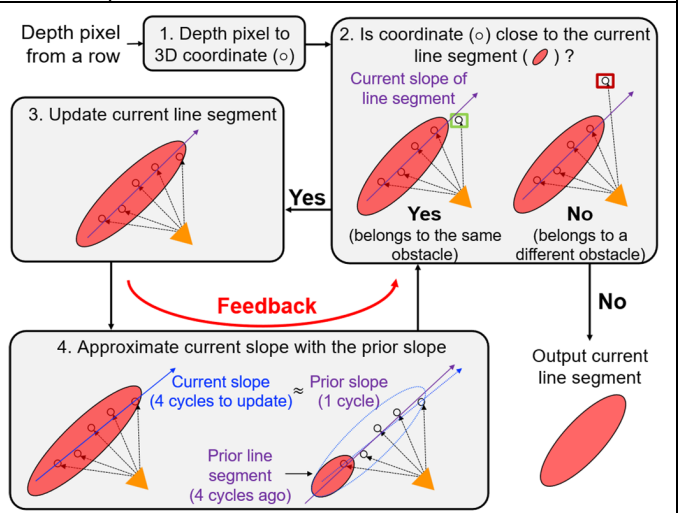

Fig. 6. A signal flow diagram of the Scanline Segmentation Unit that processes pixel by pixel. At Stage 4, the current slope requires four cycles to update. Instead of waiting for the update, the prior slope four cycles ago is retrieved in 1 cycle and feedback to Stage 2.

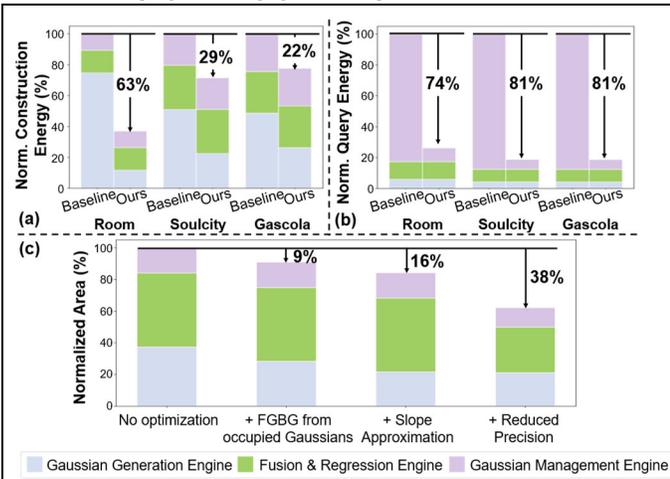

Fig. 7. Energy / area reductions of our accelerator. (a) Comparison between free Gaussian bases generation (FGBG) from line segments (baseline) and occupied Gaussians (ours). (b) Comparison between map query via one coordinate at a time (baseline) and a batch of coordinates (ours). (c) Cumulative reduction in accelerator area achieved through successive algorithm-hardware co-optimizations.

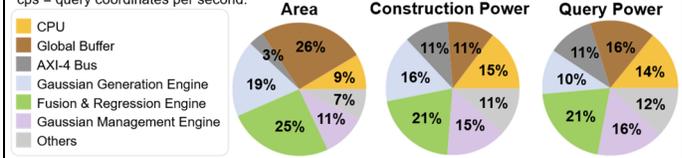

| Metrics | Environments | | | | | |
|---|---|---|---|---|---|---|
| | Room [8] (Indoor office) | | Soulcity [9] (Outdoor city) | | Gascola [9] (Outdoor forest) | |
| | Software (CPU & GPU) | Gleanmer (Ours) | Software (CPU & GPU) | Gleanmer (Ours) | Software (CPU & GPU) | Gleanmer (Ours) |
| Accuracy[1] | 96% | 96% | 99% | 99% | 96% | 96% |
| Map Size (KB) | 167 | 66 | 850 | 356 | 362 | 203 |

| Gleanmer's Performance (16 nm CMOS) | | Environments | | |
|---|---|---|---|---|
| | | Room [8] | Soulcity [9] | Gascola [9] |
| Map Construction | Throughput (fps) | 331 | 88 | 122 |
| | Power[2] (mW) | 3.7 | 4.5 | 5.3 |
| Map Query | Throughput (cps[3]) | 1320K | 540K | 876K |
| | Power[2] (mW) | 5.7 | 5.7 | 5.7 |

[1] Area under the receiver operating characteristics curve for modelling occupied and free regions [7].
[2] Peripheral overhead (I/Os, PLL) requires 38 mW (construction) to 85 mW (query) of additional idle power.
[3] cps = query coordinates per second.

Fig. 8. Top: Map accuracy and size of Gleanmer across three environments. Algorithm-hardware co-optimizations in Gleanmer do not degrade the map accuracy compared with software. Middle: Gleanmer's throughput and power consumption across all three environments. Bottom: Area and power breakdown of Gleanmer's compute engines when averaged across all three environments.

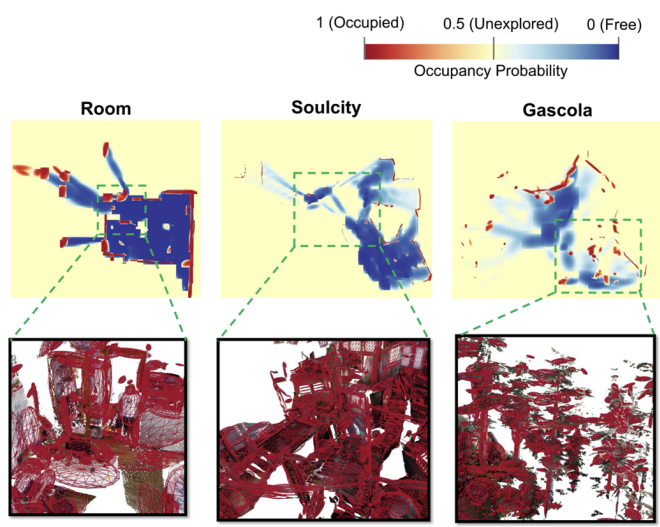

Fig. 9. Top: A 2D horizontal cross section of the 3D occupancy map where the variations of the occupancy probability are visualized from occupied (red) to unexplored (yellow) to free (blue). Bottom: Occupied Gaussians are visualized for a subregion of the 3D map.

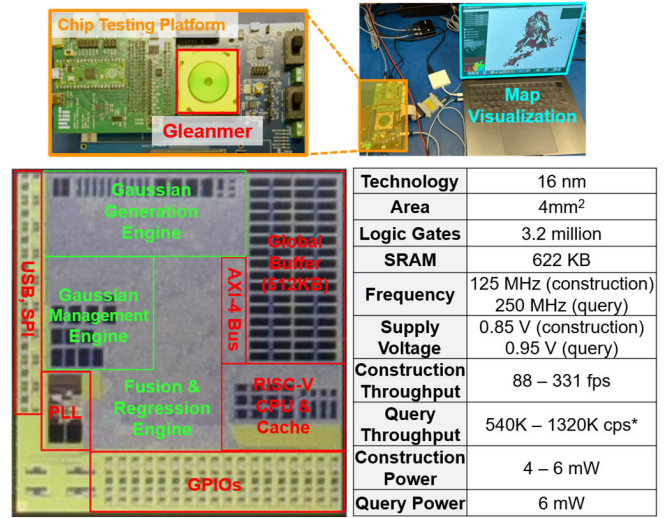

| Technology | 16 nm |
|---|---|
| Area | 4 mm² |
| Logic Gates | 3.2 million |
| SRAM | 622 KB |
| Frequency | 125 MHz (construction) 250 MHz (query) |
| Supply Voltage | 0.85 V (construction) 0.95 V (query) |
| Construction Throughput | 88 – 331 fps |
| Query Throughput | 540K – 1320K cps* |
| Construction Power | 4 – 6 mW |
| Query Power | 6 mW |

* cps = query coordinates per second

Fig. 10. Top: Chip testing and map visualization setup. Bottom: Chip micrograph with engines in the GMMap accelerator enclosed in green rectangles. Performance of Gleanmer is tabulated.

| | NVIDIA Jetson TX2 [7] | | OMU [3] | Gleanmer (Ours) |
|---|---|---|---|---|
| Technology | 16 nm | 16 nm | 12 nm | 16 nm |
| Compute Platform | ARM Cortex A57 CPU | ARM Cortex A57 CPU & 256-Core GPU | ASIC | ASIC |
| Mapping Framework | GMMap [7] | GMMap [7] | OctoMap [2] | GMMap [7] |
| Chip Area | - | - | - | 4 mm² |
| Core Area | - | - | 2.5 mm² | 2.6 mm² |
| Supply Voltage | 20 V | 20 V | 0.8 V | 0.85 V (construction) 0.95 V (query) |
| Operating Frequency | CPU: 0.4 - 2 GHz | CPU: 0.4 - 2 GHz GPU: 0.9 – 1.3 GHz | 1 GHz | 125 MHz (construction) 250 MHz (query) |
| On-chip Memory | - | - | 2 MB | 622 KB |
| Map Construction Throughput | 32 – 60 fps | 44 – 81 fps | 61 – 64 fps | 88 – 331 fps |
| Map Query Throughput | 500 – 800K cps | 500 – 800K cps | Not Reported | 540K – 1320K cps |
| Average Map Construction Power | 3.1 W | 4.7 W | 251 mW | 4.5 mW |
| Average Map Query Power | 2.0 W | 2.0 W | Not Reported | 5.7 mW |

Fig. 11. Comparison among NVIDIA Jetson TX2, OMU, and Gleanmer. With comparable accuracy (<1% difference), Gleanmer's power consumption is at least 341× lower than Jetson TX2 and 44× lower than OMU. Since GMMap is at least 5× more compact than OctoMap [7], Gleamer contains 3× less on-chip memory than OMU.